\documentclass[journal]{IEEEtran}
\usepackage{times}
\usepackage{epsfig}
\usepackage{graphicx}
\usepackage{amsmath}
\usepackage{amssymb}
\usepackage{subfigure}
\usepackage{indentfirst}
\usepackage{color}
\usepackage{multirow}
\usepackage{array}
\usepackage{booktabs}
\usepackage{algorithm}
\usepackage{algorithmic}
\usepackage{bm}
\usepackage{pifont}
\usepackage{gensymb}
\usepackage{amsmath}

\begin{document}
\title{Self-Taught Cross-Domain Few-Shot Learning with Weakly Supervised Object Localization and Task-Decomposition}

\author{Xiyao Liu,
	Zhong Ji*,
	Yanwei Pang,
	Zhongfei Zhang

\thanks{Manuscript received xxx xx, 2021; revised xxx xx, 2021.
	
	This work was supported by the National Natural Science Foundation of China (NSFC) under Grants 61771329 and 61632018.}
\thanks{X. Liu, Z. Ji*(corresponding author), and Y. Pang are with the School of Electrical and Information Engineering, Tianjin University, Tianjin 300072, China (e-mails: xiyaoliu@tju.edu.cn; jizhong@tju.edu.cn; pyw@tju.edu.cn).}
\thanks{Z. Zhang is with the Computer Science Department, Binghamton University, NY 13902, USA (e-mail: zhongfei@cs.binghamton.edu).}}

\markboth{IEEE Transactions on Multimedia,~Vol.~x, No.~x, August~xxxx}%
{Shell \MakeLowercase{\textit{et al.}}: Bare Demo of IEEEtran.cls for IEEE Journals}

\maketitle

\begin{abstract}
The domain shift between the source and target domain is the main challenge in Cross-Domain Few-Shot Learning (CD-FSL). However, the target domain is absolutely unknown during the training on the source domain, which results in lacking directed guidance for target tasks. We observe that since there are similar backgrounds in target domains, it can apply self-labeled samples as prior tasks to transfer knowledge onto target tasks. To this end, we propose a task-expansion-decomposition framework for CD-FSL, called Self-Taught (ST) approach, which alleviates the problem of non-target guidance by constructing task-oriented metric spaces. Specifically, Weakly Supervised Object Localization (WSOL) and self-supervised technologies are employed to enrich task-oriented samples by exchanging and rotating the discriminative regions, which generates a more abundant task set. Then these tasks are decomposed into several tasks to finish the task of few-shot recognition and rotation classification. It helps to transfer the source knowledge onto the target tasks and focus on discriminative regions. We conduct extensive experiments under the cross-domain setting including 8 target domains: CUB, Cars, Places, Plantae, CropDieases, EuroSAT, ISIC, and ChestX. Experimental results demonstrate that the proposed ST approach is applicable to various metric-based models, and provides promising improvements in CD-FSL.
\end{abstract}


\section{Introduction}
\IEEEPARstart{F}{ew}-Shot Learning (FSL) is to recognize novel visual concepts with several examples based on the acquired knowledge~\cite{vinyals2016matching, Snell2017Prototypical,sung2018learning}. It is also taken as a practice paradigm for meta-learning \cite{9424414,9115215}, which learns the method to transfer old knowledge on novel tasks. However, vanilla FSL methods are limited by that old and novel tasks are from the same domain. Obviously, this setting is far from the demands of real-world applications. Thus, Cross-Domain Few-Shot Learning (CD-FSL) is proposed, in which acquired knowledge and novel tasks come from the source domain and the target domain respectively. Accordingly, the domain shift from the source domain to the target one is the main challenge in CD-FSL.
\begin{figure}[t]
	\begin{center}
		\includegraphics[height=6.5cm]{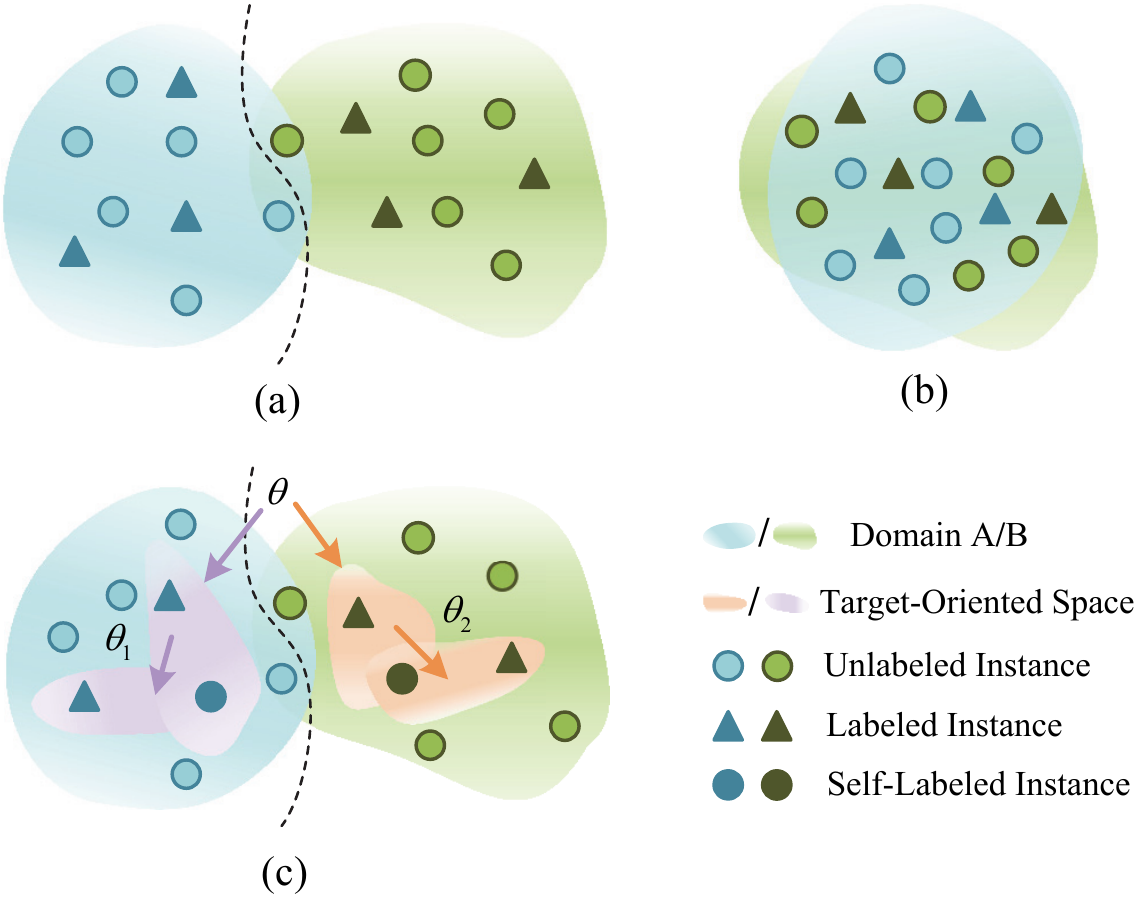}
	\end{center}
	\label{fig1}
	\caption{The illustration of Cross-Domain Few-Shot Learning. (a) Domain A and B have different distributions of embedding space. (b) Many studies aim at learning a shared space to align domain distributions. (c) Our approach applies self-labeled instances to construct target-oriented metric spaces. }
\end{figure}

There are different embedding distributions in the source and target domain, as shown in Fig. 1. The domain shift results in that the knowledge of domain A can not work well on domain B. A typical method to address this problem is aligning the two distributions by adversarial training \cite{2017Label}, domain alignment \cite{2018Boosting} and knowledge distillation \cite{2021Dynamic}. They usually choose some samples from the target domain as an auxiliary set during training. Obviously, these approaches violate the CD-FSL setting, on which the target domain is absolutely unknown during the training on the source domain. Therefore, FT \cite{2020LF} explores to realize feature transformation without access to the target domain, which employs affine transforms to simulate various feature distributions under different domains. This transformation is not directed to the target domain, which performs unstably on the different target domains. How to employ limited labeled samples to provide a directive metric space for the target source is a research focus in CD-FSL.

We observe that directly measuring the distance of the target and the source domain is difficult by limited labeled target samples. By contrast, the tasks of the same domain are highly related, so that these tasks can be decomposed to guide the metric space. This guidance can be formed by employing the labeled samples and self-labeled samples to construct multiple task-oriented metric spaces, which updates parameters to adapt for the target domain quickly, as shown in Fig.1 (c).

To construct task-oriented and robust metric spaces, it is important to generate diverse and real samples. The current setting of CD-FSL is transferring universal knowledge (miniImageNet) to fine knowledge. Since the target domains are usually fine-grained datasets, the test samples often have similar backgrounds, as shown in Fig. 2. The share of backgrounds could provide task-oriented and abundant metric spaces for task-decomposition. Thus, we introduce Weakly Supervised Object Localization (WSOL) to find the object regions.

\begin{figure}[t]
	\begin{center}
		\includegraphics[height=5.5cm]{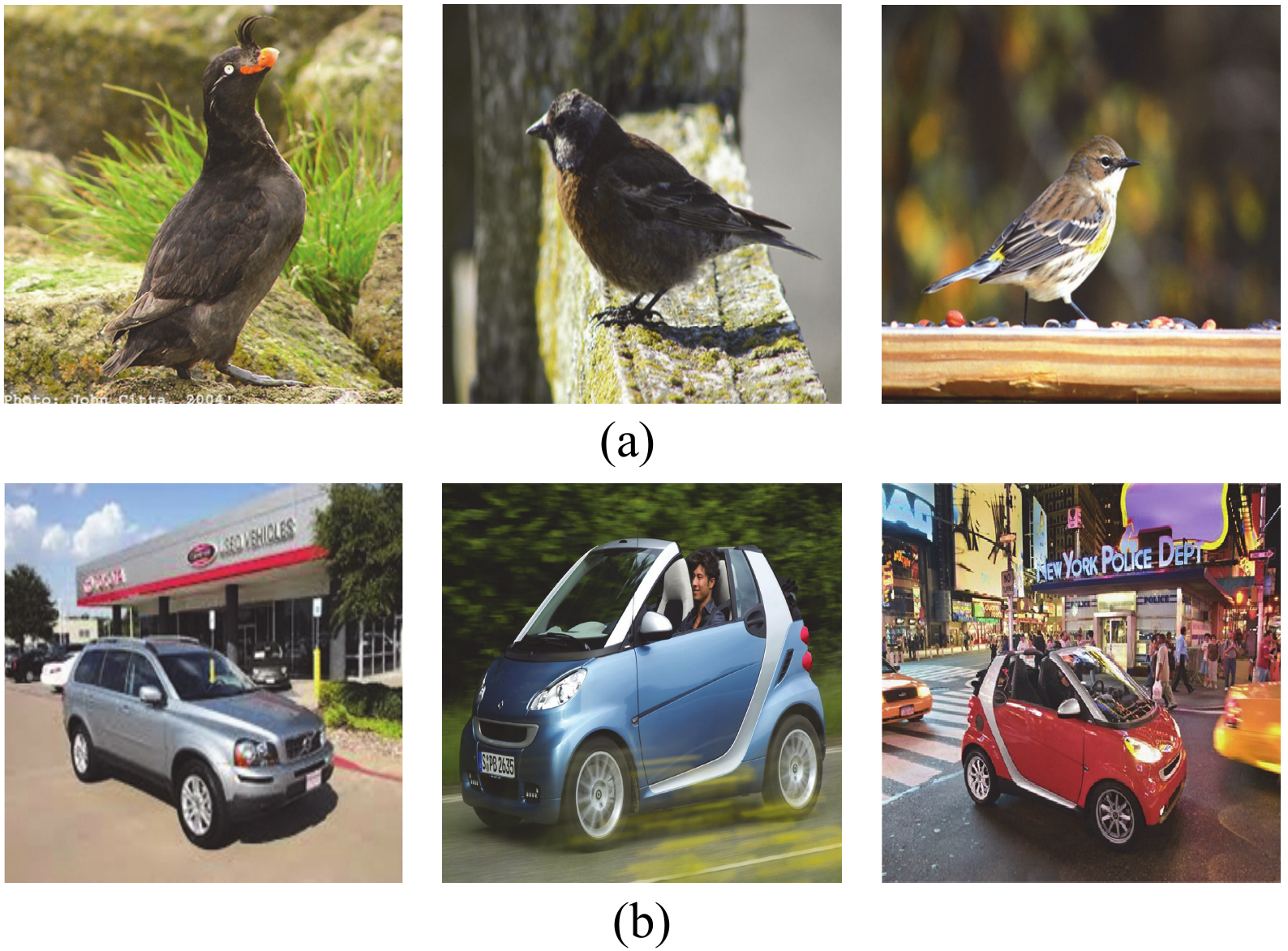}
	\end{center}
	\label{fig2}
	\caption{The examples of the target domains. There are similar and exchangeable backgrounds on the fine-grained datasets (a) CUB and (b) Cars.}
\end{figure}

WSOL aims at obtaining the localization of objects only with the image-level labels, which has broad applications in video understanding \cite{8533384}, medical imaging \cite{8674823} and remote sensing imagery analysis \cite{6915882}. Most WSOL methods extract activated regions of feature maps, which is called Class Activation Map (CAM) \cite{2016Learning}. These CAM-based approaches have achieved promising results on WSOL. Considering the capability of coverage, we apply the simplest CAM approach to find the discriminative regions and exchange the regions to realize task expansion.

To this end, we design inner tasks nesting in few-shot tasks (outer tasks), of which inner tasks learn task-oriented metric spaces by iterations. The inner tasks contain 2 stages, i.e., task-expansion and task-decomposition, which are formed by a task-expansion-decomposition framework. We name it Self-Taught (ST) approach. The illustration of ST is shown in Fig. 3. Specifically, in the Task-Expansion Module (TE-Module), weakly supervised object localization and self-supervised technologies are employed to enrich task-related samples by exchanging and rotating the discriminative regions. The TE-Module generates a more abundant task set. Then the Task-Composition Module (TD-Module) samples tasks from the set produced by TE-Module to finish the inner task, which contains the task of few-shot recognition and rotation classification. After several iterations of inner tasks, the parameters of networks are updated to the optimal metric space for the outer task, i.e., the actual few-shot task. Our ST approach is capable of fully utilizing limited labeled samples to construct target-oriented metric spaces by the task-expansion-decomposition framework.

The highlight of our work is three-fold:

1. We propose a task-expansion-decomposition framework, called Self-Taught (ST), for cross-domain few-shot learning, which alleviates the problem of undirected guidance caused by the unknown target domain. Our ST approach can work on many mainstream metric-based FSL methods, which is a plug-and-play framework.

2. We incorporate weakly supervised object localization and self-supervised technologies to construct target-oriented metric spaces, which aims at focusing on discriminative region representations. To the best of our knowledge, this is the first work to introduce weakly supervised object localization to cross-domain few-shot learning.

3. Extensive experiments on eight target benchmark datasets clearly demonstrate the effectiveness of our ST approach.

\section{Related work}
\begin{figure*}[t]
	\tiny
	\begin{center}
		\includegraphics[height=6cm,width=18cm]{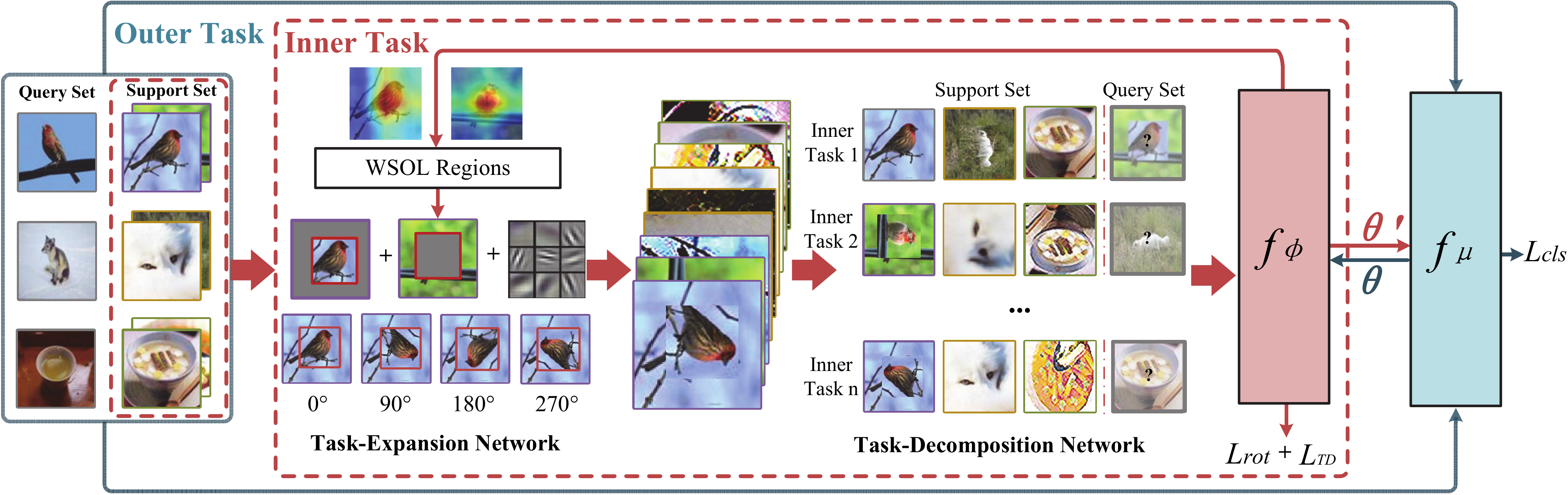}
	\end{center}
	\label{fig3}
	\caption{An overview of the proposed ST approach. The few-shot task is denoted as outer task, and its support set inputs inner task iterations. Firstly, WSOL generates the object locations from the feature maps of labeled samples. The task-expansion network applies the locations and original images to augment tasks by rotations, exchanges and random convolution filters. Then, task-decomposition network samples inner tasks to update the metric-based few-shot method $f_\phi$. When finishing the inner tasks, the outer task is trained to update the metric-based few-shot method $f_\mu$.
	}
\end{figure*} 

\subsection{Few-Shot Learning}

The task of Few-Shot Learning (FSL) is proposed to recognize new categories only from few samples for scarce annotations. The current mainstream FSL methods are built on the meta-learning framework, which transfers prior knowledge to novel tasks. FSL can be mainly categorized into 3 approaches: optimization-based, metric-based, and generation-based methods. 

Optimization-based approaches are designed to generalize to novel tasks by a few steps of updates. MAML~\cite{finn2017model-agnostic} employs a second-order optimizing strategy with the meta-learning framework by carrying out a small number of update steps. Then Nichol $ et~al. $~\cite{nichol2018reptile} simplified the implementation of MAML only with the first-order derivation and presented Reptile to move meta-learner parameters towards base-learner parameters. Follow this line, many variants \cite{Rusu2018Meta,9157767,2018Recasting,8954011} have made incremental changes for MAML. For example, LEO \cite{Rusu2018Meta} applies a low-dimensional latent space to update parameters, which is more practical in low-data scenarios. Since the shared initialization may lead to conflict over tasks, Baik $et\ al$. \cite{9157767} proposed a task-and-layer-wise attenuation to forget the prior information selectively.

Metric-based approaches aim at constructing an embedding space to measure similarities between the labeled and unlabeled images. This line of approaches mainly depends on designing a suitable embedding space and a distance measurement method, which is simple and effective. Matching Network~\cite{vinyals2016matching} combines attention mechanism with LSTM to obtain the probability distribution belonging to the categories in the support set. Besides, it also proposes the episode training paradigm, which is widely employed in FSL. Relation Network~\cite{sung2018learning} employs deep convolution networks to replace the artificial distance function, which achieves promising performance. Snell $ et~al. $~\cite{Snell2017Prototypical} assumed that there are class prototypes in the embedding space for each class. They designed Prototypical Network, which averages the samples in the support set as class prototypes. The above three approaches have established the foundation of metric-based FSL. Many studies conduct incremental improvements based on them, such as \cite{oreshkin2018tadam,9311850,8578527,8953758,9091240}.

The main idea of the generation-based FSL methods improves the generalizability of the classifier by producing diverse samples. Generative Adversarial Network (GAN) is a typical generator in FSL. For example, Wang $et\ al$. \cite{8578858} applied GAN to generate samples, which improves performance by about $6\%$. And Li $et\ al$. \cite{9156726} employed cWGAN to ensure the diversity of generated data. Besides, some studies employ auxiliary approaches to directly generate image-level samples. For example, Zhang $et\ al$. \cite{8953363} introduced saliency-based masks to exchange foregrounds and backgrounds of different images to augment samples.

\subsection{Cross-Domain Few-Shot Learning}
Facing realistic FSL application, it is almost impossible that a large number of labeled samples and unlabeled test samples come from the same domain. It usually occurs that there are abundant auxiliary samples in related domains. Therefore, Cross-Domain Few-Shot Learning (CD-FSL) is proposed, where labeled and unlabeled samples belong to the source domain and the target domain respectively. The domain shift problem caused by different distributions is a challenge for CD-FSL.

A typical method to address the domain shift problem is aligning the two distributions by adversarial training \cite{2017Label}, domain alignment \cite{2018Boosting} and knowledge distillation \cite{2021Dynamic}. Besides, Chen $et\ al$. \cite{chen2019closer} conducted experiments employing several popular vanilla FSL algorithms on the setting of CD-FSL, which demonstrates that the pre-trained and fine-tuned approaches could improve the performance on the unseen domain. Although these approaches achieve promising results, they applied labeled or unlabeled samples from the target domain before testing. Therefore, they break the CD-FSL setting that the target domain is absolutely unknown during the training.

By contrast, some work explores to improve the knowledge transferability without access to the target domain. For example, FT \cite{2020LF} employs the affine transformation to simulate various feature distributions under different domains. Since this transformation is not directed to the target domain, it performs unstably on the different target domains. LRP \cite{2020LRP} applies the model-agnostic explanation-guided training strategy to improve the feature representations. ATA \cite{2021ATA} considers the task distribution space under the worst-case problem by adversarial task augmentation, which strengthens the robustness of the inductive bias. Different from them, we focus on how to employ limited labeled samples to construct metric spaces pointing to the target source. We propose a task-expansion-decomposition framework for CD-FSL, which obtains optimal parameters by target-oriented inner tasks.

\subsection{Weakly Supervised Object Localization}
Weakly Supervised Object Localization (WSOL) obtains the localization of objects only with the image-level labels. Most WSOL methods extract activated regions of feature maps, which is called Class Activation Map (CAM) \cite{2016Learning}. It employs a global average pooling (GAP) layer to preserve the spatial information of the feature maps and extracts the spatial region of the feature responses from the last convolutional feature map. Later, Grad-CAM \cite{2020Grad} is proposed to simplify CAM, which extracts the heatmap from any convolutional feature map.

Although CAM-based methods are simple and effective, they probably focus on the broad regions of objects without the marginal parts. Thus, an erasing-based strategy is designed to preserve more relevant parts of objects. According to the erasing regions, they are categorized into two types: random erasing \cite{8237643,9008296} and network-guided erasing methods \cite{9156910}. Apart from the above erasing methods, SPG \cite{2018Self} and I$^2$C \cite{2020Inter} apply the constraint of pixel-level correlations to improve the quality of localization maps.

In recent years, weakly supervised object localization techniques have been employed in video understanding \cite{8533384}, medical imaging \cite{8674823} and remote sensing imagery analysis \cite{6915882}. In our work, we apply weakly supervised object localization as prior knowledge to generate task-oriented samples and focus on the discriminative regions, which provides task transferring space in cross-domain few-shot learning.

\section{Self-Taught Approach}
\subsection{Problem Formulation}
We follow the common episodic paradigm in \cite{vinyals2016matching,Snell2017Prototypical,sung2018learning}, which is trained on the $N$-way $K$-shot setting. There are $K$ labeled samples and a number of unlabeled samples from $N$ categories composing of a few-shot task. The support set consists of $N \times K$ labeled samples, denoted as $S= \{x_i^s,y_i^s\}_{i=1}^{N\times K}$. And the query set is represented as $Q= \{x_i^q,y_i^q\}_{i=1}^{M}$, where $M$ is the total number of unlabeled samples. In vanilla FSL, the train set and test set belong to the same domain. While in CD-FSL, the train set and test set are sampled from different domains.

\renewcommand{\algorithmicrequire}{\textbf{Input:}}  
\renewcommand{\algorithmicensure}{\textbf{Output:}} 
\begin{algorithm}[t]
	\caption{The training algorithm of the Self-Taught approach on $N$-way $K$-shot.} 
	\label{alg1}
	\begin{algorithmic}[1]
		\REQUIRE The train set $ \bm{D_{train}}$, the task settings including the number of category $K$, the number of support examples per class $N$ and the total number of query examples $ M $. 
		\ENSURE Learned model.
		\STATE Randomly initialize parameters
		\WHILE{training}
		\STATE $\bm{T} = TaskSample(\bm{D_{train}},N,K,M)$
		\FOR{$ t $ in $ \left\{\bm{T}\right\} $}
			\STATE $ \{(x_i^s,y_i^s)\}^{NK}\in \bm{S}, \{(x_i^q,y_i^q)\}^M\in \bm{Q}$
			\STATE $Loc_i = WSOL(x_i^s)$
			\STATE $Back_i = x_i^s- Loc_i$
			\STATE $\hat{x_i^s},\hat{y_i^\omega} = rot(Loc_i) \oplus Back_j $
			\STATE $ \{\hat{x_i^s}, y_i^s ,\hat{y_i^\omega}\} \in \bm{S'}$
			\STATE $\bm{T^{inner}} = TaskSample(\bm{S'},N,K,M')$
			\FOR{$ t^{inner} $ in $ \left\{\bm{T^{inner}}\right\} $}
				\STATE $ \{(x_i^{\widetilde{s}},y_i^{\widetilde{s}},y_i^{\widetilde{s}\omega})\}^{NK}\in \bm{\widetilde{S}}, \{(x_i^{\widetilde{q}},y_i^{\widetilde{q}})\}^{M'}\in \bm{\widetilde{Q}}$
				\STATE $y_i^{\omega}=E(f_\phi(x_i^{\widetilde{s}}))$
				\STATE $ c_{k} =   {\dfrac{1}{N}\sum\limits f_\phi(\hat{x_i^s})} $
				\FOR{$ (x_i^{\widetilde{q}},y_i^{\widetilde{q}}) $ in $  \bm{\widetilde{Q}} $ }
					\STATE $ Loss(t^{inner}|\theta')=-\dfrac{1}{NK}\sum\limits y_i^{\widetilde{s}\omega} log(y_i^{\omega})+ \dfrac{1}{M'}[ log \sum\limits_{w'} exp(-d(f_\phi(x_i^{\widetilde{s}}),c_{k'}))+d( f_\phi(x_i^{\widetilde{s}}), c_{k} )] $
				\ENDFOR 		
			\ENDFOR 
			\STATE $ {c_{k}}={\dfrac{1}{N}\sum\limits f_\mu(x_i^s)} $
			\FOR{$ (x_i^q,y_i^q) $ in $  \bm{Q} $ }
				\STATE $ Loss(t|\theta)=\dfrac{1}{M}[log \sum\limits_{w'} exp(-d(f_\mu(x_i^q),c_{k'}))+d( f_\mu(x_i^q), c_{k} )] $ 
			\ENDFOR
		\ENDFOR
		\ENDWHILE	
	\end{algorithmic}
\end{algorithm}

\subsection{Overview}
In this work, we propose a Self-Taught (ST) approach for Cross-Domain Few-Shot Learning (CD-FSL) employing Weakly Supervised Object Localization (WSOL) and task-decomposition, which is shown in Fig. 3. Our ST approach consists of a Task-Expansion Module (TE-Module) and a Task-Decomposition Module (TD-Module). The TE-Module applies WSOL and self-supervised technologies to generate novel samples, which enables networks focus on discriminative regions and further provides task-oriented spaces for task-decomposition. Then the TD-Module samples few-shot tasks from novel labeled set to update network parameters. This task-expansion-decomposition paradigm learns to adjust parameters according to novel tasks from different domains.

The training procedure is shown in Alg. 1. There are 2 types of tasks in our proposed ST approach. The support set inputs inner tasks to update the inner network, and the traditional few-shot task is trained as the outer task. There are the same metric-based few-shot methods in the outer task and inner task but in different parameter update ways. Our ST approach can work on metric-based few-shot methods, i.e., MatchingNet \cite{vinyals2016matching}, ProtoNet \cite{Snell2017Prototypical}, RelationNet \cite{sung2018learning}. Although graph-based methods are superior in performance, they can not deal with the varying number of query samples by the same network. Therefore, our ST approach is mainly designed on the metric-based approaches. We introduce our method by taking ProtoNet as an example, where $f_\phi$ and $f_\mu$ are the inner network and outer network with the same architecture and initial parameters.

\subsection{Task-Expansion Module}
The Task-Expansion Module (TE-Module) employs WSOL and self-supervised technologies to generate novel labeled samples and learn more discriminative features. Different from other self-supervised approaches, we apply WSOL to obtain the attentive locations of images, which utilizes the focus region to exchange and rotate:
\begin{equation}
Loc_i = WSOL(x_i^s),
\end{equation}
where $x_i^s$ is the $i$-th sample in support set and $Loc_i$ is the object region of $x_i^s$. We apply a simple WSOL approach \cite{2016Learning} to obtain the location of objects, which is shown in Fig. 4. Firstly, feature maps are extracted from the images in the support set by the feature extractor in metric-based methods, which can be denoted as:
\begin{equation}
p_i = f_\phi(x_i^s),
\end{equation}
where $f_\phi$ is the CNN encoder of ProtoNet in the inner task, and $p_i$ is the feature map of $x_i^s$. Then we employ Class Activation Map (CAM) \cite{2016Learning} to highlight the discriminative regions, which is represented as:
\begin{equation}
h_i = CAM(p_i),
\end{equation}
where $h_i$ is the heatmap of $p_i$. To obtain bounding boxes from $h_i$, we employ an auto-thresholding technique to segment the heatmap. We turn the heatmap to a binary image by the threshold. Besides, we take the square bounding box that covers the largest connected component in the binary image:
\begin{equation}
Loc_i = Seg(h_i).
\end{equation}
Accroding to the object location, the image can be divided into two parts:
\begin{equation}
Back_i = x_i^s- Loc_i,
\end{equation}
where $Back_i$ is the background of $x_i^s$. Our ST approach requires generating novel task-oriented samples for task-decomposition. Thus, we design 2 types of generating methods: 1) Rotating the object localization. Different from the rotation of the whole image, the network can focus on the object location, especially for small object boxes. 2) Exchanging and randomly rotating foregrounds and backgrounds. Since there are similar backgrounds in fine-grained datasets, the exchanges of inter-class and intra-class provide diverse task-oriented samples. They are defined as:
\begin{equation}
\hat{x_i^s},\hat{y_i^\omega} =
\begin{cases}
rot(Loc_i) \oplus Back_i,& i=j\\
rot(Loc_i) \oplus Back_j,& i\neq j
\end{cases} 
\end{equation}
where $rot(\ )$ rotates the $Loc_i$ by the random angle from $ [0\degree, 90\degree, 180\degree, 270\degree]$, $\hat{y_i^\omega}$ is the angle label from $[0,1,2,3]$, $\oplus$ represents the image stitching and random convolutions, $\hat{x_i^s}$ is the novel sample. Random convolutions aim at smoothing images and adding noise. Thus, the expanded support set $S'=\{\hat{x_i^s}, y_i^s ,\hat{y_i^\omega}\}$ contains more samples.

\begin{figure}[t]
	\begin{center}
		\includegraphics[width=8.8cm]{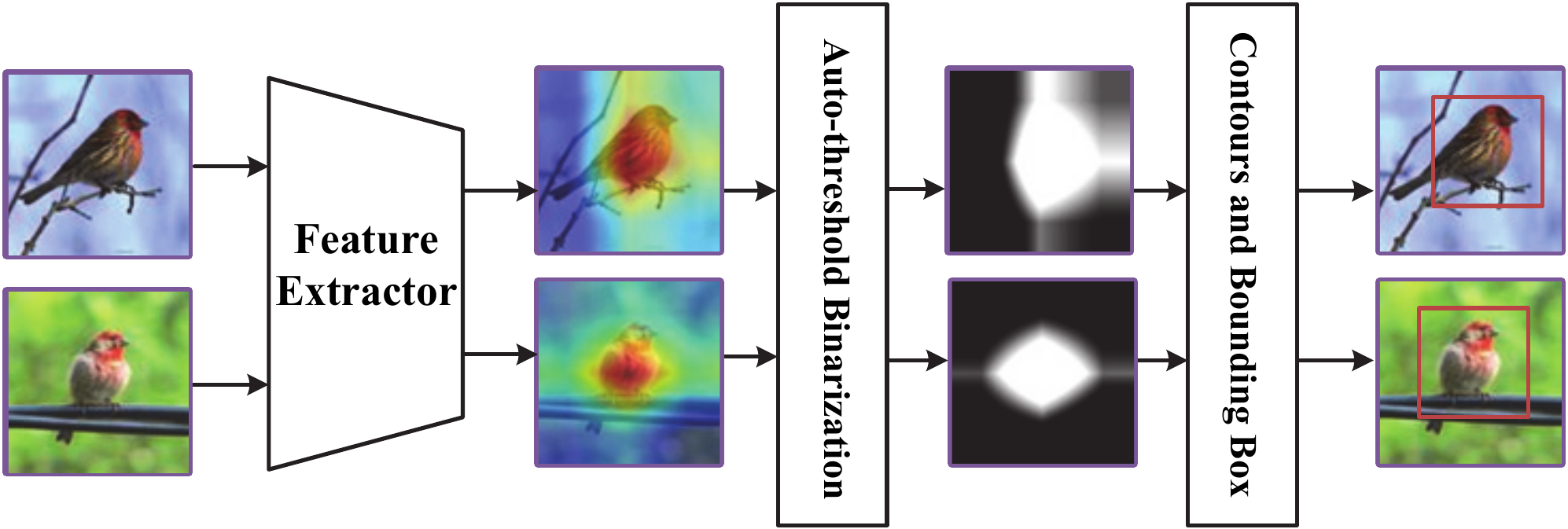}
	\end{center}
	\label{fig4}
	\caption{An illustration of the WSOL method. It extracts feature maps and employs CAM to obtain heatmaps. Then the auto-thresholding technique is employed to binarize the heatmap and the max bounding box is found to cover the object regions.}
\end{figure}

\subsection{Task-Decomposition Module}
Task-Decomposition Module (TD-Module) samples $\alpha$ iterations of $N$-way $K$-shot task from $S'$, including the support set $\widetilde{S} =\{(x_i^{\widetilde{s}},y_i^{\widetilde{s}},y_i^{\widetilde{s}\omega})\}^{NK}$ and the query set $\widetilde{Q} = \{(x_i^{\widetilde{q}},y_i^{\widetilde{q}})\}^{M'}$.

The inner task is a multi-task framework, which includes the rotation classification and few-shot recognition. For the rotation classification, a nonlinear network $E$ embeds features to its rotation angle label as:
\begin{equation}
y_i^{\omega}=E(f_\phi(x_i^{\widetilde{s}})),
\end{equation}
where $y_i^{\omega}$ is the predicted rotation label. We apply a cross-entropy loss to measure the rotation angle error, which is defined as:
\begin{equation}
Loss_{rot}=-\dfrac{1}{NK}\sum\limits y_i^{\widetilde{s}\omega} log(y_i^{\omega}).
\end{equation}

For the inner few-shot recognition, the class prototypes are calculated by:
\begin{equation}
 c_{k} =   {\dfrac{1}{K}\sum\limits f_\phi(\hat{x_i^s})},
\end{equation}
where $c_k$ is the class prototypes of the $k$-th category. The encoder $f_\phi$ gets as input an sample $x_i^s$ with its few-shot label $y_i^s$, and yields as output a probability distribution over all possible classification scores:
\begin{equation}
{p(y=j|x_i^{\widetilde{q}})} =  \dfrac{exp(-d(x_i^{\widetilde{q}},c_{j}))}{\sum_{K}exp(-d(x_i^{\widetilde{q}},c_{k}))},
\end{equation}
where $d(\ )$ is Euclidean Distance. The few-shot classification loss is:
\begin{equation}
Loss_{TD}= \dfrac{1}{M'}[ log \sum\limits_{w'} exp(-d(f_\phi(x_i^{\widetilde{s}}),c_{k'}))+d( f_\phi(x_i^{\widetilde{s}}), c_{k} )].
\end{equation}

The inner loss consists of rotation angle prediction loss and few-shot classification loss, which is denoted as:
\begin{equation}
Loss(t^{inner}|\theta')=Loss_{TD}+\lambda Loss_{rot},
\end{equation}
where $\lambda$ is a hyper-paremeter to balance the two tasks, and $\theta'$ is parameters of the inner network.

After $\alpha$ iterations of inner tasks, the outer network $f_\mu$ is updated on $\theta'$. Similarly, the class prototypes and the classification scores are obtained by:
\begin{equation}
{c_{k}} =  {\dfrac{1}{k}\sum\limits f_\mu(x_i^s)},
\end{equation}

\begin{equation}
{p(y=j|x_i^{q})} =  \dfrac{exp(-d(x_i^{q},c_{j}))}{\sum_{K}exp(-d(x_i^{q},c_{k}))}.
\end{equation}

The outer network $f_\mu$ applies a cross-entropy loss to measure the classification error, which is denoted as:
\begin{small}
	\begin{equation}
Loss_{cls}=\dfrac{1}{M}[log \sum\limits_{w'} exp(-d(f_\mu(x_i^q),c_{k'}))
+d( f_\mu(x_i^q), c_{k} )].
	\end{equation}
\end{small}

\section{Experimental Setup}
We conduct experiments to prove the effectiveness of our proposed ST approach on three popular metric-based methods: MatchingNet \cite{vinyals2016matching}, ProtoNet \cite{Snell2017Prototypical}, and RelationNet \cite{sung2018learning}. Under cross-domain settings, the source domain is miniImageNet \cite{ravi2017optimization}, and the target domains contain 8 fine-grained datasets.

\begin{table}[h]
	\label{Table.1}
	\caption{\upshape Summarization of the datasets.}
	\centering
	\renewcommand{\arraystretch}{1.2}
	\setlength{\tabcolsep}{8pt}
	\begin{tabular}{cccc}
		\hline
		Dataset      & Train Set               & Validation Set & Test Set  \\ \hline
		miniImageNet & 64                         & 16             & 20        \\
		CUB          & 100                        & 50             & 50        \\
		Cars         & 98                         & 49             & 49        \\
		Places       & 183 & 91             & 91        \\
		Plantae      & 100                        & 50             & 50        \\
		CropDieases  & 38                           &   -             &   38        \\
		EuroSAT      & 10                           &   -             &  10         \\
		ISIC         &7                & -     & 7 \\
		ChestX       &   14                         &      -          &    14       \\ \hline
	\end{tabular}
\end{table}

\begin{table*}[]
	\label{Table.2}
	\caption{\upshape  Cross-domain few-shot classification accuracy on test splits of CUB, Cars, Places and Plantae datasets with $\pm$ 95\% confidence intervals.}
	\centering
	\renewcommand{\arraystretch}{1.2}
	\setlength{\tabcolsep}{3pt}
	\scalebox{0.9}{\begin{tabular}{ccccccccc}
			\hline
			& \multicolumn{2}{c}{CUB}                                           & \multicolumn{2}{c}{Cars}                            & \multicolumn{2}{c}{Places}                                                      & \multicolumn{2}{c}{Plantea}                                       \\
			& 5-way 1-shot                           & 5-way 5-shot             & 5-way 1-shot             & 5-way 5-shot             & 5-way 1-shot                           & 5-way 5-shot                           & 5-way 1-shot                           & 5-way 5-shot             \\ \hline
			MatchingNet \cite{vinyals2016matching}                      & 41.46\% $\pm$ 0.4\%                        & 61.21\% $\pm$ 0.4\%          & 31.82\% $\pm$ 0.3\%          & 48.11\% $\pm$ 0.4\%          & 48.83\% $\pm$ 0.5\%                        & 66.29\% $\pm$ 0.4\%                        & 36.56\% $\pm$ 0.3\%                        & 53.81\% $\pm$ 0.3\%          \\
			+FT \cite{2020LF}                              & 43.59\% $\pm$ 0.4\%                        & 61.65\% $\pm$ 0.4\%          & 35.03\% $\pm$ 0.3\%          & 47.72\% $\pm$ 0.4\%          & \textbf{53.00\% $\pm$ 0.5\%}               & 66.01\% $\pm$ 0.4\%                        & 36.28\% $\pm$ 0.3\%                        & 51.12\% $\pm$ 0.3\%          \\
			+ATA \cite{2021ATA}                             & 41.59\% $\pm$ 0.4\%                        & 59.33\% $\pm$ 0.4\%          & 35.14\% $\pm$ 0.3\%          & 48.78\% $\pm$ 0.4\%          & 51.86\% $\pm$ 0.5\%          & 66.31\% $\pm$ 0.4\%                        & 37.02\% $\pm$ 0.3\%                        & 51.56\% $\pm$ 0.3\%          \\
			\textbf{+ST (Ours)} & \textbf{46.22\% $\pm$ 0.4\%}               & \textbf{73.87\% $\pm$ 0.4\%} & \textbf{35.29\% $\pm$ 0.3\%} & \textbf{60.29\% $\pm$ 0.4\%} & {52.75\% $\pm$ 0.5\%} & \textbf{69.56\% $\pm$ 0.4\%}               & \textbf{37.25\% $\pm$ 0.3\%}               & \textbf{61.30\% $\pm$ 0.3\%} \\ \hline
			ProtoNet \cite{Snell2017Prototypical}                          & 41.42\% $\pm$ 0.4\%                        & 62.69\% $\pm$ 0.4\%          & 31.26\% $\pm$ 0.3\%          & 48.35\% $\pm$ 0.4\%          & 48.44\% $\pm$ 0.5\%                        & 70.47\% $\pm$ 0.4\%                        & 32.17\% $\pm$ 0.3\%                        & 53.13\% $\pm$ 0.3\%          \\
			+FT \cite{2020LF}                              & 41.74\% $\pm$ 0.4\%                        & 63.84\% $\pm$ 0.4\%          & 31.47\% $\pm$ 0.3\%          & 48.98\% $\pm$ 0.3\%          & 48.41\% $\pm$ 0.5\%          & 69.20\% $\pm$ 0.4\%                   & 33.02\% $\pm$ 0.3\%                        & 52.31\% $\pm$ 0.3\%          \\
			+ATA \cite{2021ATA}                        & 39.85\% $\pm$ 0.4\%                        & 64.55\% $\pm$ 0.4\%          & 32.21\% $\pm$ 0.3\%          & 50.44\% $\pm$ 0.4\%          & 49.15\% $\pm$ 0.5\%                        & 69.94\% $\pm$ 0.4\%                        & 31.72\% $\pm$ 0.3\%                        & 53.14\% $\pm$ 0.3\%          \\
			\textbf{+ST (Ours)} & \textbf{43.97\% $\pm$ 0.4\%}               & \textbf{74.62\% $\pm$ 0.4\%} & \textbf{33.74\% $\pm$ 0.3\%} & \textbf{59.38\% $\pm$ 0.4\%} & \textbf{50.31\% $\pm$ 0.5\%}               & \textbf{73.71\% $\pm$ 0.4\%}               & \textbf{33.96\% $\pm$ 0.3\%}               & \textbf{62.72\% $\pm$ 0.3\%} \\ \hline
			RelationNet \cite{sung2018learning}                       & 41.27\% $\pm$ 0.4\%                        & 56.77\% $\pm$ 0.4\%          & 30.09\% $\pm$ 0.3\%          & 40.46\% $\pm$ 0.4\%          & 48.16\% $\pm$ 0.5\%                        & 64.25\% $\pm$ 0.4\%                        & 31.23\% $\pm$ 0.3\%                        & 42.71\% $\pm$ 0.3\%          \\
			+FT \cite{2020LF}                              & \textbf{43.33\% $\pm$ 0.4\%}               & 59.77\% $\pm$ 0.4\%          & 30.45\% $\pm$ 0.3\%          & 40.18\% $\pm$ 0.4\%          & 49.92\% $\pm$ 0.5\%                        & 65.55\% $\pm$ 0.4\%                        & 32.57\% $\pm$ 0.3\%                        & 44.29\% $\pm$ 0.3\%          \\
			+LRP \cite{2020LRP}                             & 41.57\% $\pm$ 0.4\%                        & 57.70\% $\pm$ 0.4\%          & 30.48\% $\pm$ 0.3\%          & 41.21\% $\pm$ 0.4\%          & 48.47\% $\pm$ 0.5\%                        & 65.35\% $\pm$ 0.4\%                        & 32.11\% $\pm$ 0.3\%                        & 43.70\% $\pm$ 0.3\%          \\
			+ATA \cite{2021ATA}                             & 43.02\% $\pm$ 0.4\%                        & 59.36\% $\pm$ 0.4\%          & 31.79\% $\pm$ 0.3\%          & 42.95\% $\pm$ 0.4\%          & \textbf{51.16\% $\pm$ 0.5\%}               & \textbf{66.90\% $\pm$ 0.4\%}               & \textbf{33.72\% $\pm$ 0.3\%}               & 45.32\% $\pm$ 0.3\%          \\
			\textbf{+ST (Ours)} & {43.10\% $\pm$ 0.4\%} & \textbf{62.94\% $\pm$ 0.4\%} & \textbf{32.34\% $\pm$ 0.3\%} & \textbf{43.26\% $\pm$ 0.4\%} & {50.53\% $\pm$ 0.5\%} & {66.74\% $\pm$ 0.4\%} & {33.19\% $\pm$ 0.3\%} & \textbf{46.92\% $\pm$ 0.3\%} \\ \hline
	\end{tabular}}
\end{table*}

\begin{table*}[]
	\label{Table.3}
	\caption{\upshape Cross-domain few-shot classification accuracy on test splits of CropDieases, EuroSAT, ISIC and ChestX datasets with $\pm$ 95\% confidence intervals.}
	\centering
	\renewcommand{\arraystretch}{1.2}
	\setlength{\tabcolsep}{3pt}
	\scalebox{0.9}{\begin{tabular}{ccccccccc}
			\hline
			& \multicolumn{2}{c}{CropDieases}                                            & \multicolumn{2}{c}{EuroSAT}                          & \multicolumn{2}{c}{ISIC}                                                                          & \multicolumn{2}{c}{ChestX}                                                 \\
			& 5-way 1-shot                                    & 5-way 5-shot             & 5-way 1-shot              & 5-way 5-shot             & 5-way 1-shot                                    & 5-way 5-shot                                    & 5-way 1-shot                                    & 5-way 5-shot             \\ \hline
			MatchingNet \cite{vinyals2016matching} & 61.49\% $\pm$ 0.4\%                                 & 85.12\% $\pm$ 0.4\%          & 58.30\% $\pm$ 0.3\%           & 77.54\% $\pm$ 0.4\%          & 32.79\% $\pm$ 0.3\%                                 & 45.12\% $\pm$ 0.3\%                                 & 22.25\% $\pm$ 0.2\%                                 & 25.52\% $\pm$ 0.2\%          \\
			+FT \cite{2020LF}         & 64.95\% $\pm$ 0.5\%                                 & 83.39\% $\pm$ 0.4\%          & 57.56\% $\pm$ 0.3\%           & 72.49\% $\pm$ 0.3\%          & 32.73\% $\pm$ 0.3\%                                 & 43.44\% $\pm$ 0.3\%                                 & 22.76\% $\pm$ 0.2\%                                 & 25.13\% $\pm$ 0.2\%          \\
			+ATA \cite{2021ATA}        & 66.09\% $\pm$ 0.4\%                        & 86.36\% $\pm$ 0.4\%          & 59.65\% $\pm$ 0.3\%           & 76.96\% $\pm$ 0.4\%          & 33.34\% $\pm$ 0.3\%                                 & 42.53\% $\pm$ 0.3\%                                 & 22.74\% $\pm$ 0.2\%                                 & 25.11\% $\pm$ 0.2\%          \\
			\textbf{+ST (Ours)} & \textbf{ 66.65\% $\pm$ 0.4\%}          & \textbf{90.86\% $\pm$ 0.4\%} & \textbf{64.96\% $\pm$ 0.3\%}  & \textbf{86.83\% $\pm$ 0.4\%} & { \textbf{34.12\% $\pm$ 0.3\%}} & \textbf{59.99\% $\pm$ 0.3\%}                        & \textbf{22.85\% $\pm$ 0.2\%}                        & \textbf{27.01\% $\pm$ 0.2\%} \\ \hline
			ProtoNet \cite{Snell2017Prototypical}    & 60.23\% $\pm$ 0.4\%                                 & 86.83\% $\pm$ 0.4\%          & 56.41\% $\pm$ 0.3\%           & 78.67\% $\pm$ 0.4\%          & 31.21\% $\pm$ 0.3\%                                 & 45.89\% $\pm$ 0.3\%                                 & 22.10\% $\pm$ 0.2\%                                 & 26.24\% $\pm$ 0.2\%          \\
			+FT  \cite{2020LF}        & 60.68\% $\pm$ 0.4\%                                 & 85.95\% $\pm$ 0.4\%          & 56.02\% $\pm$ 0.3\%           & 78.49\% $\pm$ 0.4\%          & 29.80\% $\pm$ 0.3\%                                 & 44.91\% $\pm$ 0.3\%                                 & 21.90\% $\pm$ 0.2\%                                 & 26.14\% $\pm$ 0.2\%          \\
			+ATA \cite{2021ATA}        & 61.25\% $\pm$ 0.4\%                                 & 88.29\% $\pm$ 0.4\%          & 55.45\% $\pm$ 0.3\%           & 77.93\% $\pm$ 0.4\%          & 30.85\% $\pm$ 0.3\%                                 & 45.32\% $\pm$ 0.3\%                                 & 22.26\% $\pm$ 0.2\%                                 & 26.09\% $\pm$ 0.2\%          \\
			\textbf{+ST (Ours)} & \textbf{63.53\% $\pm$ 0.4\%}                        & \textbf{92.40\% $\pm$ 0.4\%} & \textbf{62.62\% $\pm$ 0.3\%}  & \textbf{88.35\% $\pm$ 0.4\%} & \textbf{31.92\% $\pm$ 0.2\%}                       & \textbf{61.55\% $\pm$ 0.3\%}                        & \textbf{22.46\% $\pm$ 0.2\%}                        & \textbf{28.40\% $\pm$ 0.2\%} \\ \hline
			RelationNet \cite{sung2018learning} & 53.58\% $\pm$ 0.4\%                                 & 72.86\% $\pm$ 0.4\%          & 49.08\% $\pm$ 0.4\%           & 65.56\% $\pm$ 0.4\%          & 30.53\% $\pm$ 0.3\%                                 & 38.60\% $\pm$ 0.3\%                                 & 21.95\% $\pm$ 0.2\%                                 & 24.07\% $\pm$ 0.2\%          \\
			+FT \cite{2020LF}         & 57.57\% $\pm$ 0.5\%                                 & 75.78\% $\pm$ 0.4\%          & 53.53\% $\pm$ 0.4\%           & 69.13\% $\pm$ 0.4\%          & 30.38\% $\pm$ 0.3\%                                 & 38.68\% $\pm$ 0.3\%                                 & 21.79\% $\pm$ 0.2\%                                 & 23.95\% $\pm$ 0.2\%          \\
			+LRP \cite{2020LRP}        & 55.01\% $\pm$ 0.4\%                                 & 74.21\% $\pm$ 0.4\%          & 50.99\% $\pm$ 0.4\%           & 67.54\% $\pm$ 0.4\%          & 31.16\% $\pm$ 0.3\%                                 & 39.97\% $\pm$ 0.3\%                                 & 22.11\% $\pm$ 0.2\%                                 & 24.28\% $\pm$ 0.2\%          \\
			+ATA \cite{2021ATA}        & 61.17\% $\pm$ 0.4\%                                 & 78.20\% $\pm$ 0.4\%          & 55.69\% $\pm$ 0.4\%           & 71.02\% $\pm$ 0.4\%          & 31.13\% $\pm$ 0.3\%                                 & 40.38\% $\pm$ 0.3\%                                 & 22.14\% $\pm$ 0.2\%                                 & 24.43\% $\pm$ 0.2\%          \\
			\textbf{+ST (Ours)} & {\textbf{63.29\% $\pm$ 0.4\%}} & \textbf{78.62\% $\pm$ 0.4\%} & \textbf{57.36\% $\pm$ 0.30\%} & \textbf{75.84\% $\pm$ 0.4\%} & {\textbf{32.09\% $\pm$ 0.3\%}} & {\textbf{44.42\% $\pm$ 0.3\%}} & { \textbf{22.28\% $\pm$ 0.2\%}} & \textbf{24.79\% $\pm$ 0.2\%} \\ \hline
	\end{tabular}}
\end{table*}

\subsection{Datasets}
There are 9 datasets involving in our experiment, including 1 source domain and 8 target domains. Their information is shown in Table \uppercase\expandafter{\romannumeral1}.

\textbf{miniImageNet.} miniImageNet is proposed by \cite{ravi2017optimization}, which is a subset of ImageNet \cite{2009ImageNet}. It contains 100 classes with 600 images per class, which covers animals, food, tools, furniture and etc. Due to its universality, it is usually employed as the source domain in CD-FSL. In our work, we follow the split in \cite{ravi2017optimization}, in which we apply train and validation set for training.

\textbf{CUB.} CUB-200 (CUB) \cite{2011CUB} is a benchmark dataset for fine-grained bird recognition, including 200 classes with a total of 11,788 images. We take its test set as one of the target domains, following the split in \cite{2018Few}.

\textbf{Cars.} Stanford Cars (Cars) \cite{20143D} is proposed for fine-grained car classification. It contains 16,185 images of 196 classes of cars, whose labels are typically at the level of Make, Model, and Year.

\textbf{Places.} Places365-Standard (Places) \cite{Places} is a benchmark dataset for the scene understanding task, which totally contains 1,803,460 images within 365 scene categories. The number of images per class varies from 3,068 to 5,000.

\textbf{Plantae.} This dataset is extracted from \cite{Plantae}, which contains 200 species of plants. Since the natural world is heavily imbalanced, the number of each class is different but reaches at least 100 samples.

\textbf{CropDieases.} This dataset \cite{2016Using} is a public dataset of 54,306 diseased and healthy plant samples, labeling 14 crop species and 26 diseases. The version in CD-FSL is reorganized with the format ``crop species\_disease'', which contains 38 classes.
	
\textbf{EuroSAT.} It is a novel dataset based on Sentinel-2 satellite images \cite{EuroSAT}, which covers 13 spectral bands with a total of 27,000 labeled and geo-referenced images. And it contains 10 classes, including annual crop, permanent crop, pastures, highway, residential buildings, industrial buildings, and etc. 

\textbf{ISIC.} ISIC2018 (ISIC) \cite{2018ISIC} consists of 10,015 dermatoscopic images and is publicly available through the ISIC archive. This benchmark dataset is employed for comparisons with machine learning and human experts.

\textbf{ChestX.} ChestX-ray8 (ChestX) \cite{ChestX} is a new chest X-ray database, which comprises 108,948 frontal view X-ray images of 32,717 unique patients with 14 types of disease pattern.

Note that the splits of CropDieases, EuroSAT, ISIC, and ChestX we process are exactly the same as \cite{2021ATA,2020Abroader}.

\subsection{Experimental Settings}
As for the visual representations, we apply the pretrained ResNet-$10 $~\cite{he2016deep} to extract the visual features. The hyperparameter $\lambda$ is set to $ 0.1 $. For the outer task, we employ Adam optimizer with initial learning rate with $ 0.0001 $. And for the inner task, we apply SGD optimizer with learning rate with $ 0.001 $. $\lambda$ of the inner loss is 0.1. We utilize the most popular settings in few-shot learning: $ 5 $-way $ 1 $-shot and $ 5 $-way $ 5 $-shot settings. The number of query set is 15 per class.

\subsection{Competitors}

Several FSL models are chosen as competitors: 

(1) Metric-based approaches, including Matching Networks \cite{vinyals2016matching}, Prototypical Networks \cite{Snell2017Prototypical}, and Relation Networks \cite{sung2018learning}. 

(2) Cross-domain few-shot approaches, including FT \cite{2020LF}, ATA \cite{2021ATA} and LRP \cite{2020LRP}. They can work on the above metric-based approaches. While LRP only implements on RelationNet due to its requiring of classifier parameters.

To fairly compare with the approaches, we utilize ResNet-10 \cite{he2016deep} as embedding networks for all the methods. The recognition accuracies for CD-FSL are computed by averaging over $ 2000 $ randomly generated episodes.

\begin{table*}[th]
	\label{Table.4}
	\caption{\upshape Ablation studies of the ST approach on CUB and Places datasets.}
	\centering
	\renewcommand{\arraystretch}{1.2}
	\setlength{\tabcolsep}{8pt}
	\begin{tabular}{cccccccc}
		\hline
		\multicolumn{2}{c}{WSOL}                              & Whole                & \multicolumn{1}{l}{} & \multicolumn{2}{c}{CUB}                             & \multicolumn{2}{c}{Places}                          \\
		\multicolumn{1}{c}{Rot} & \multicolumn{1}{c}{Exc-Rot} & Rot                  & TD                   & 5-way 1-shot             & 5-way 5-shot             & 5-way 1-shot             & 5-way 5-shot             \\ \hline
		&                             & \multicolumn{1}{l}{} & \multicolumn{1}{l}{} & 41.42\% $\pm$ 0.4\%          & 62.69\% $\pm$ 0.4\%          & 48.44\% $\pm$ 0.5\%          & 70.47\% $\pm$ 0.4\%          \\
		\multicolumn{1}{c}{$ \checkmark $}   &                             & \multicolumn{1}{l}{} & \multicolumn{1}{l}{} & 42.78\% $\pm$ 0.4\%          & 66.77\% $\pm$ 0.4\%          & 49.72\% $\pm$ 0.5\%          & 71.49\% $\pm$ 0.4\%          \\
		& \multicolumn{1}{c}{$ \checkmark $}       & \multicolumn{1}{l}{} & \multicolumn{1}{l}{} & 42.96\% $\pm$ 0.4\%          & 67.35\% $\pm$ 0.4\%          & 49.67\% $\pm$ 0.5\%          & 71.56\% $\pm$ 0.4\%          \\
		&                             & $ \checkmark $                    & \multicolumn{1}{l}{} & 42.76\% $\pm$ 0.4\%          & 65.89\% $\pm$ 0.4\%          & 49.69\% $\pm$ 0.5\%          & 71.53\% $\pm$ 0.4\%          \\
		&                             & \multicolumn{1}{l}{} & $\checkmark $                    & 43.16\% $\pm$ 0.4\%          & 70.46\% $\pm$ 0.4\%          & 50.07\% $\pm$ 0.5\%          & 72.79\% $\pm$ 0.4\%          \\
		\multicolumn{1}{c}{$ \checkmark $}   &                             &                      & $ \checkmark $                    & 43.34\% $\pm$ 0.4\%          & 72.21\% $\pm$ 0.4\%          & 50.19\% $\pm$ 0.5\%          & 73.32\% $\pm$ 0.4\%          \\
		& \multicolumn{1}{c}{$ \checkmark $}       &                      & $ \checkmark $                    & 43.49\% $\pm$ 0.4\%          & 73.65\% $\pm$ 0.4\%          & 50.23\% $\pm$ 0.5\%          & 73.39\% $\pm$ 0.4\%          \\
		&                             & $ \checkmark $                    & $ \checkmark $                    & 43.33\% $\pm$ 0.4\%          & 71.58\% $\pm$ 0.4\%          & 50.20\% $\pm$ 0.5\%          & 73.25\% $\pm$ 0.4\%          \\
		\multicolumn{1}{c}{$ \checkmark $}   & \multicolumn{1}{c}{$ \checkmark $}       & \multicolumn{1}{l}{} & $ \checkmark $                    & \textbf{43.97\% $\pm$ 0.4\%} & \textbf{74.62\% $\pm$ 0.4\%} & \textbf{50.31\% $\pm$ 0.5\%} & \textbf{73.71\% $\pm$ 0.4\%} \\ \hline
	\end{tabular}
\end{table*}

\subsection{Comparison with State-of-the-Art Methods}

We conduct experiments on the most common setting in FSL, $ 5 $-way $ 1 $-shot and $ 5 $-way $ 5 $-shot on all datasets. We apply the train and validation set of miniImageNet for training and then test the model on eight target domains.

The results on CUB, Cars, Places, and Plantae are shown in Table \uppercase\expandafter{\romannumeral2}. It can be observed that the proposed ST approach achieves significant improvements over most approaches on both $ 5 $-way $ 1 $-shot and $ 5 $-way $ 5 $-shot settings. Specifically, the proposed model achieves accuracies of $ 73.87\% $ (MatchingNet + ST) and $ 74.62\% $ (ProtoNet + ST) on CUB, $ 60.29\% $ (MatchingNet + ST) and $ 59.38\% $ (ProtoNet + ST) on Cars under $ 5 $-way $ 5 $-shot setting, outperforming other approaches in a large margin. We notice that ATA demonstrates competitive performance on RelationNet, which obtains the accuracy of $ 51.16\% $ and $ 69.90\% $ on Places under both settings. While the performance of our ST approach slightly declines on Places dataset, we infer that the WSOL loses its superior for scene classification. Since there is no specific object location in the scene, it turns into a normal self-supervised approach. We employ ablation studies to prove this inference, which is shown in Section IV.E.

Comparing with the above datasets, Cropdieases, EuroSAT, ISIC, ChestX datasets are more natural and realistic. The results are shown in Table \uppercase\expandafter{\romannumeral3}. It is observed that the proposed ST approach clearly beats all competitive approaches on all datasets. Besides, we observe that the most significant gains appear on $5$-way $5$-shot by ProtoNet + ST. It achieves accuracy gains of $ 5.56\% $, $ 9.68\% $, $ 15.66\% $, and $ 2.16\% $ respectively against ProtoNet on the four datasets. Especially, the ChestX dataset is the hardest dataset for CD-FSL due to the huge difference between the source domain (miniImageNet) and the target domain, on which other approaches are almost ineffective. By contrast, our ST approach obtains satisfactory performance of $ 27.01\% $ (MatchingNet + ST), $ 28.40\% $ (ProtoNet + ST), and $ 24.79\% $ (RelationNet + ST) on ChestX. It demonstrates that our task-expansion-decomposition framework is capable of narrowing domain-shift. Our ST approach employs weakly supervised object localization and self-supervised approaches to generate task-oriented samples and then implements inner tasks to learn a task-specific metric space. Besides, it utilizes a way of exchanging and rotating to make the network focus on discriminative regions and learn more robust visual information. Therefore, ST approach has a clear improvement against the competitors.
		
\begin{figure}[htb]
	\centering                                                                              
	\begin{subfigure}[The impacts of different iterations of the inner task on $ 5 $-way $ 1 $-shot.]
		{
			\includegraphics[height=4.5cm]{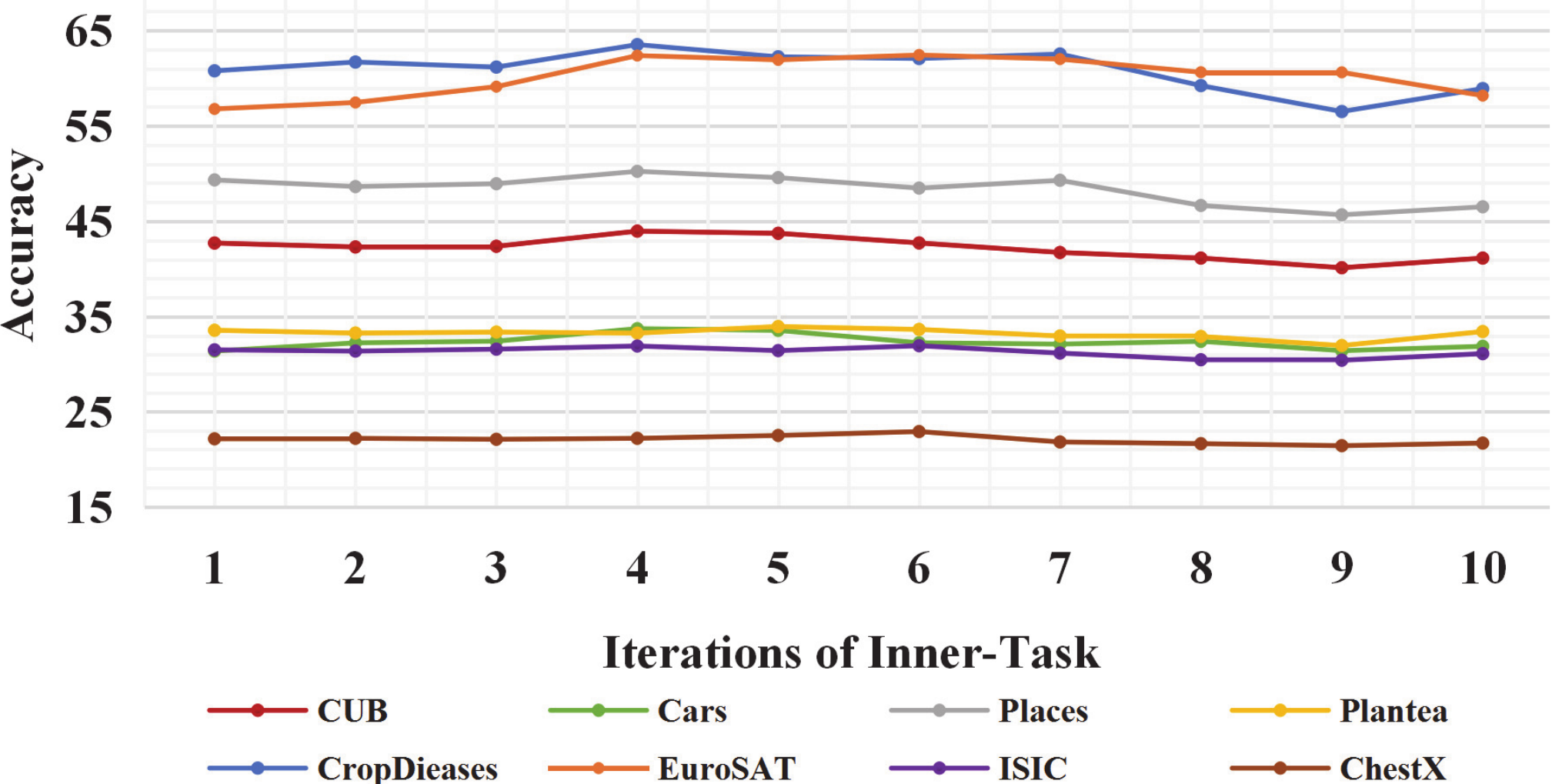}
		}
	\end{subfigure}
	\begin{subfigure}[The impacts of different iterations of the inner task on $ 5 $-way $ 5 $-shot.]
		{
			\includegraphics[height=4.6cm]{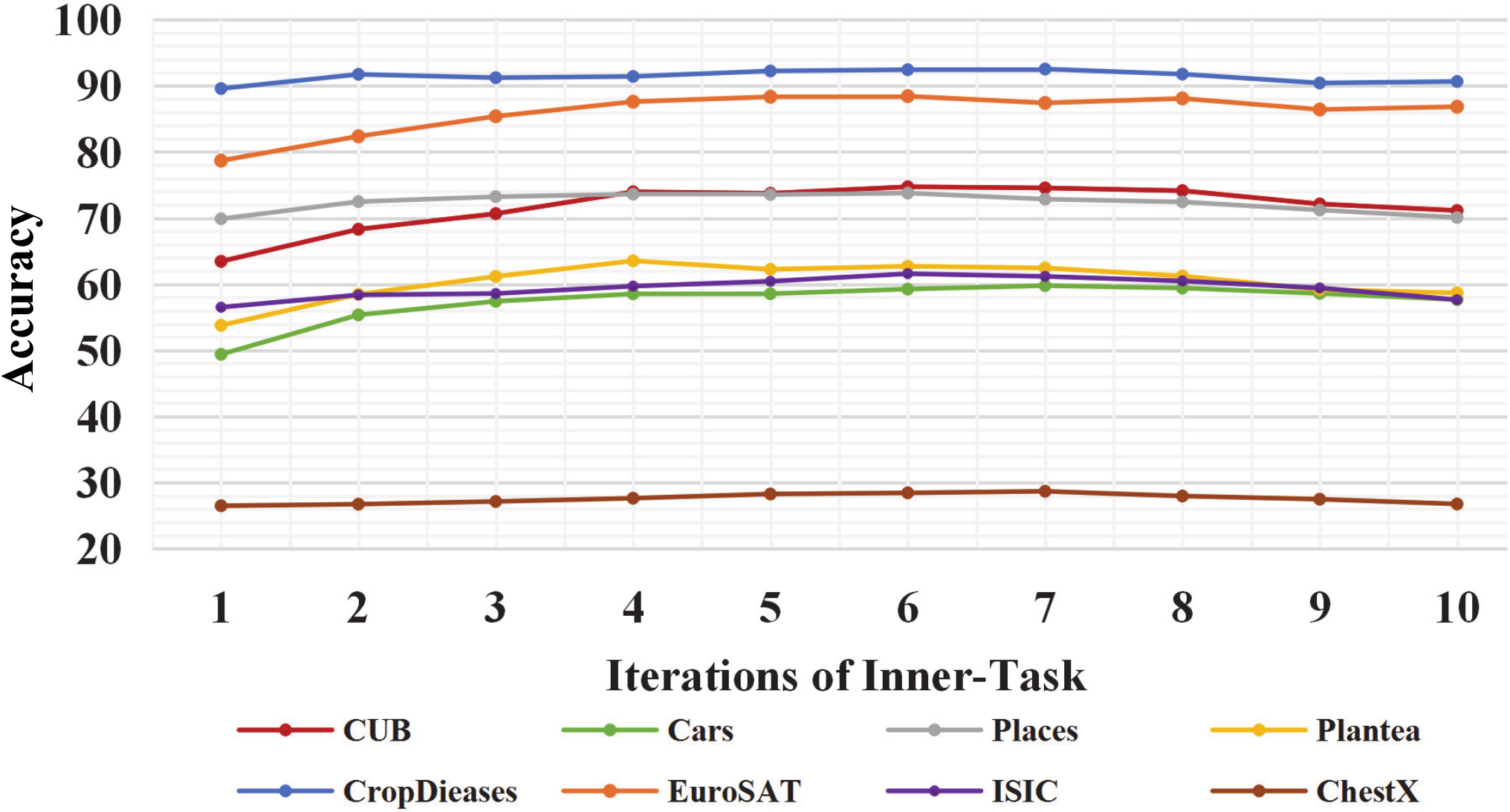}
		}
	\end{subfigure}	
	
	\caption{\upshape{The influences of varied iterations of the inner task $ \alpha $ on all datasets. }}
	\label{fig5}
\end{figure}

\subsection{Ablation Studies}

In order to evaluate the effectiveness of each component, we conduct experiments as ablation studies. We choose two datasets, i.e., CUB and Places, according to that whether there exist concrete objects in the images.

The components are designed as follows:

WSOL Rot: Rotating the region obtained by WSOL to generate samples.

WSOL Exc-Rot: Exchanging and rotating the region obtained by WSOL to generate samples.

Whole Rot: Rotating the whole images to generate samples.

TD: Task-decomposition.

\begin{figure*}[t]
	\tiny
	\begin{center}
		\includegraphics[height=8.5cm]{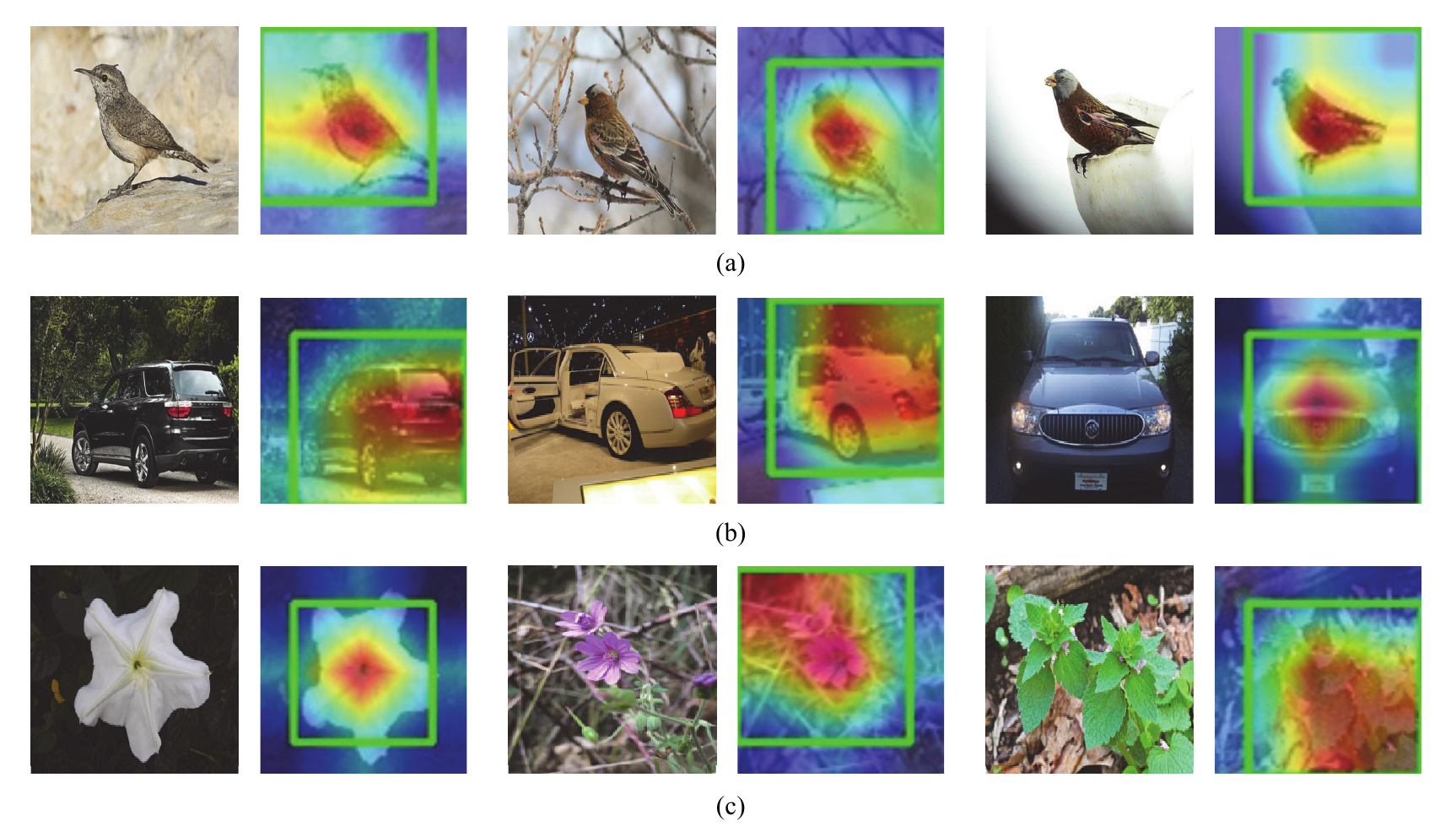}
	\end{center}
	\label{fig6}
	\caption{Examples of weakly supervised object location results on (a) CUB, (b) Cars and (c) Plantae datasets.
	}
\end{figure*} 
\begin{figure*}[t]
	\tiny
	\begin{center}
		\includegraphics[height=5cm]{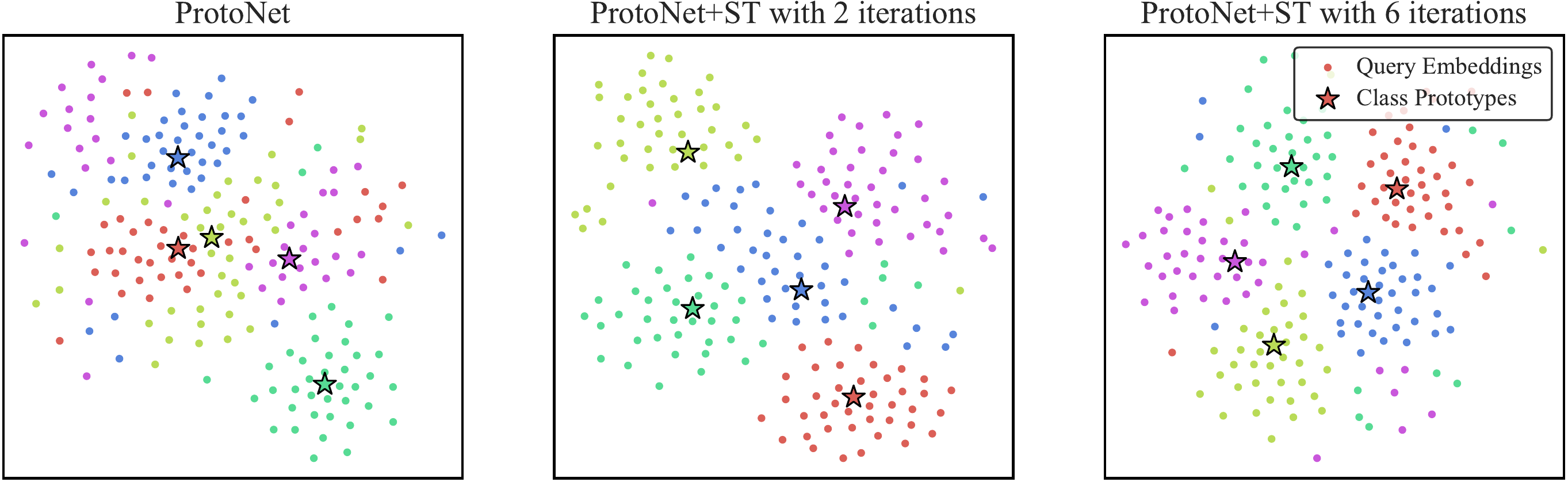}
	\end{center}
	\label{fig7}
	\caption{The t-SNE visualization of the embeddings learned by ProtoNet, ProtoNet+ST with 2 iterations, ProtoNet+ST with 6 iterations on CropDieases.
	}
\end{figure*}

The experimental results of different components are shown in Table \uppercase\expandafter{\romannumeral4}. We have the following observations: 1) TD has significant impacts on results, which brings about $ 1.74\% $ and $ 7.77\% $ performance gains on CUB and $ 1.63\% $ and $ 2.05\% $ performance gains on Places under $ 5 $-way $ 1 $-shot and $ 5 $-way $ 5 $-shot settings respectively. 2) On CUB, WSOL Exc-Rot performs better than the other two approaches (WSOL Rot and Whole Rot) at most $ 0.2\% $ and $ 2.07\% $ on $ 5 $-way $ 1 $-shot and $ 5 $-way $ 5 $-shot. 3) On Places, three generation approaches, WSOL Exc-Rot, WSOL Rot, and Whole Rot have similar performance. It proves that WSOL does not work well on the scene classification without obvious objects. It also can be observed that any one generation approach + TD is capable of bringing performance gains on both datasets. The both WSOL approaches + TD achieve the best results, which makes up the proposed ST approach.

\subsection{The impact of iterations $\alpha$}		
The quantitative analysis for iterations of the inner task $\alpha$ is illustrated in Fig. 5 on the eight target domains under both $ 5 $-way $ 1 $-shot and $ 5 $-way $ 5 $-shot settings. The hyper-parameter $ \alpha$ is applied for adjusting the strength of task decomposition. We can observe that there are similar trends on both settings. Specifically, on $ 5 $-way $ 1 $-shot, the performance gradually grows with an earlier increase of $ \alpha $, which illustrates that the iterations of the inner task $\alpha$ make contributions to cross-domain few-shot recognition. However, the peak performance of the different target domains is obtained by different $\alpha$, probably from 4 to 6. Considering the trade-off between cost and performance, we choose $ \alpha =4 $ as the iterations of the inner task on $ 5 $-way $ 1 $-shot. Then, the performance begins to decrease slightly. We infer that excessive iterations hurt the generalization performance for the next task. Similarly, for the $ 5 $-way $ 5 $-shot setting, the optimal value of $\alpha$ is 6$\sim$8. Due to the same reason, we select $ \alpha =6 $.

\subsection{Qualitative Results}
\subsubsection{WSOL on unseen classes}					
We visualize WSOL results on CUB, Cars, and Plantae datasets to evaluate the impacts on unseen classes, which is shown in Fig. 6. It can be observed that although the feature extractor does not have access to the target domains, it still performs robustly. When the main object appears in the image, WSOL gives the approximate location boxes on three datasets. Since the target domains are all fine-grained datasets, they have similar backgrounds. We design the Exc-Rot based on WSOL to generate task-oriented samples by exchanging backgrounds and objects between inner-class and intra-class. Experimental results have proved that WSOL plays an important role in our approach.

\subsubsection{Visualization Analysis}
To prove the effectiveness of our method, we visualize the embedding space on the CropDieases dataset by the t-SNE approach, which is shown in Fig. 7. It can be observed that the class prototypes of ProtoNet obtained by averaging labeled samples deviate from some query samples, which extremely hurts the classification performance for CD-FSL. We choose iterations of the inner task $\alpha = 2$ and $\alpha = 6$ to verify our ST approach. Obviously, the embeddings obtained from ProtoNet+ST with 2 iterations are more separable than ProtoNet. Moreover, ProtoNet+ST with 6 iterations further improve the performance, where the shapes of class clusters are similar and balanced. The t-SNE visualization demonstrates our ST approach is capable of addressing the cross-domain few-shot task.

\section{Conclusion}			
This paper has introduced a Self-Taught (ST) approach to alleviate the problem of non-directed guidance for target tasks in cross-domain few-shot learning. Our ST approach can nest in various metric-based approaches and is a task-expansion-decomposition framework. It first employs weakly supervised object localization and self-supervision technologies to generate fine-grained level samples. Then they decompose into subtasks to finish the task of few-shot recognition and rotation classification. It not only constructs task-oriented metric spaces but also focuses on the discriminative regions. The experimental results show the impressive effectiveness of our ST approach on eight target domains.

\bibliographystyle{ieeetr}
\bibliography{egbib}

\begin{IEEEbiography}[{\includegraphics[width=1in,height=1.25in,clip,keepaspectratio]{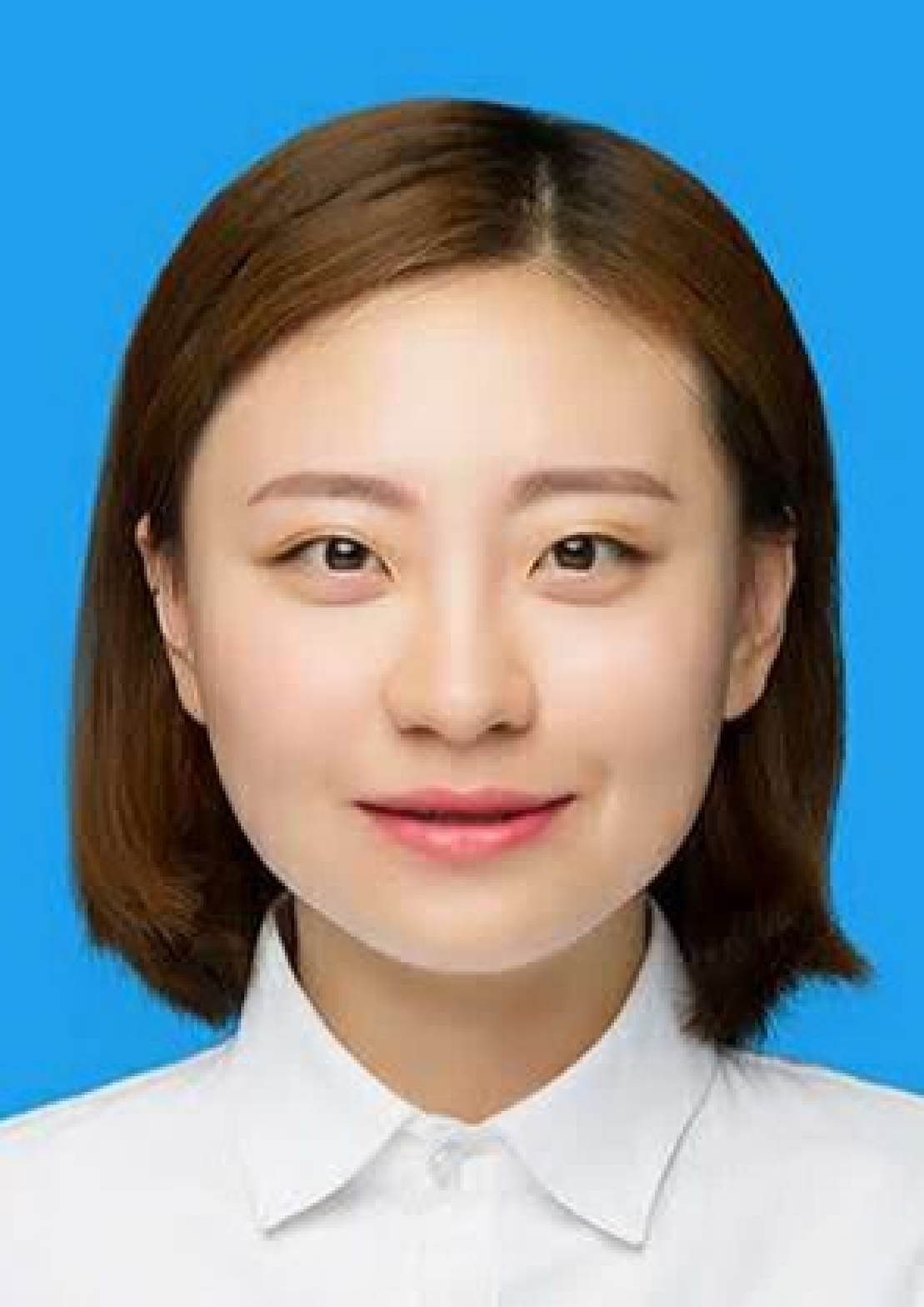}}]{Xiyao Liu}
	received the B.S. degree in telecommunication engineering from Tianjin University, Tianjin, China, in 2015. She is currently pursuing a Ph.D. degree in the School of Electrical and Information Engineering, Tianjin University. Her research interests include few-shot learning, human-object interaction, and computer vision.
\end{IEEEbiography}

\begin{IEEEbiography}[{\includegraphics[width=1in,height=1.25in,clip,keepaspectratio]{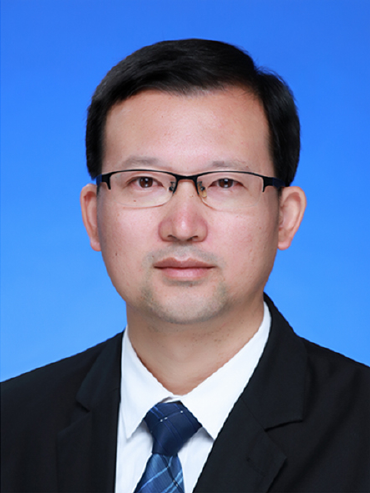}}]{Zhong Ji}
	received the Ph.D. degree in signal and information processing from Tianjin University, Tianjin, China, in 2008. He is currently a Professor with the School of Electrical and Information Engineering, Tianjin University. He has authored over 100 scientific papers. His current research interests include multimedia understanding, zero/few-shot leanring, cross-modal analysis, and video summarization.
\end{IEEEbiography}

\begin{IEEEbiography}[{\includegraphics[width=1in,height=1.25in,clip,keepaspectratio]{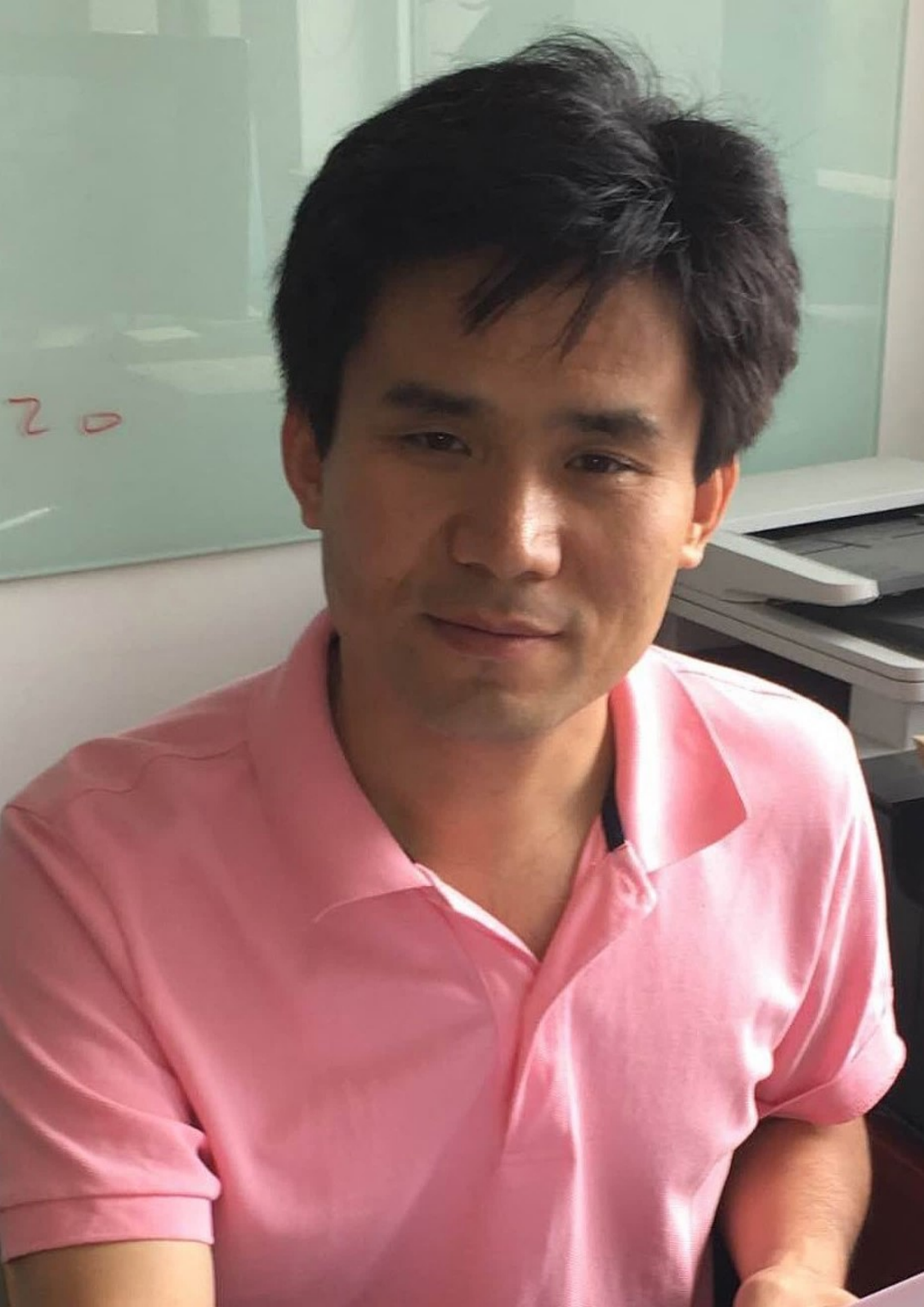}}]{Yanwei Pang}
	received the Ph.D. degree in electronic engineering from the University of Science and Technology of China, Hefei, China, in 2004. He is currently a Professor with the School of Electrical and Information Engineering, Tianjin University, Tianjin, China. He has authored over 120 scientific papers. His current research interests include object detection and recognition, vision in bad weather, and computer vision.
\end{IEEEbiography}

\begin{IEEEbiography}[{\includegraphics[width=1in,height=1.25in,clip,keepaspectratio]{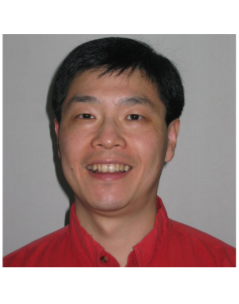}}]{Zhongfei Zhang}
	received the BS degree in electronics engineering (with honors), the MS degree in information science, both from Zhejiang University, Hangzhou, China, and the PhD degree in computer science from the University of Massachusetts at Amherst. He is currently a professor at the Computer Science Department and directs the Multimedia Research Laboratory in the State University of New York (SUNY) at Binghamton, USA. He has published more than 200 peer-reviewed academic papers.
\end{IEEEbiography}

\end{document}